\documentclass[cameraready]{Interspeech}

\title{Learning task-specific subspaces via interventional post-training of speech foundation models}

\author[affiliation={1}, orcid=0009-0002-0892-5764, correspondingauthor]{Jack}{Cox}
\author[affiliation={1}, orcid=0000-0002-1684-5660]{Jon P}{Barker}

\address{
    $^1$ University of Sheffield, United Kingdom
}

\email{jbjcox1@sheffield.ac.uk}


\usepackage{comment}
\usepackage[printonlyused]{acronym}
\usepackage{tikz}
\usepackage{multirow}
\usepackage{nicematrix}
\usepackage{hyperref}

\usetikzlibrary{shapes,decorations,arrows,calc,arrows.meta,fit,positioning,shapes.geometric}

\acrodef{scm}[SCM]{Structural Causal Model}
\acrodef{ssl}[SSL]{Self-Supervised Learning}
\acrodef{asr}[ASR]{Automatic Speech Recognition}
\acrodef{ica}[ICA]{Independent Component Analysis}
\acrodef{tts}[TTS]{Text-to-Speech}
\acrodef{sv}[SV]{Speaker Verification}
\acrodef{awe}[AWE]{Acoustic Word Embedding}
\acrodef{eer}[EER]{Equal Error Rate}
\acrodef{sid}[SID]{Speaker Identification}
\acrodef{mlp}[MLP]{Multi-Layer Perceptron}
\acrodef{ood}[OOD]{Out-of-Domain}
\acrodef{ic}[IC]{Intent Classification}
\acrodef{acc}[acc]{accuracy}
\acrodef{ks}[KS]{Keyword Spotting}

\def\Secref#1{Section~\ref{#1}}

\def\eqref#1{equation~\ref{#1}}

\def\1{\bm{1}}

\def\va{{\bm{a}}}
\def\vb{{\bm{b}}}

\def\vh{{\bm{h}}}

\def\vp{{\bm{p}}}
\def\vq{{\bm{q}}}

\def\vx{{\bm{x}}}

\def\vz{{\bm{z}}}

\def\mH{{\bm{H}}}
\def\mI{{\bm{I}}}
\def\mJ{{\bm{J}}}

\def\mL{{\bm{L}}}
\def\mM{{\bm{M}}}

\def\mZ{{\bm{Z}}}

\DeclareMathAlphabet{\mathsfit}{\encodingdefault}{\sfdefault}{m}{sl}
\SetMathAlphabet{\mathsfit}{bold}{\encodingdefault}{\sfdefault}{bx}{n}

\def\emM{{M}}

\begin{document}

\maketitle

\begin{abstract}
    Speech foundation models, pre-trained on large corpora of unlabelled speech data, produce general-purpose representations which are useful across tasks. However, these representations encode information about salient speech variables in a distributed manner, while downstream speech tasks rely on only some of this variability. In this work, we propose a post-training refinement approach using interventional contrastive learning. By leveraging an interventional dataset and multi-part contrastive loss, we learn a transformation from the entangled representation space of speech foundation models into separate content and speaker subspaces. We evaluate the learnt representations on speaker verification and keyword spotting tasks, showing improved out-of-domain speaker verification performance and evidence that speaker and content information are separated across the learned subspaces.
\end{abstract}

\section{Introduction} 

In recent years, \ac{ssl} has emerged as an important paradigm for speech tasks, leveraging large quantities of unlabelled data to train general-purpose representation models \cite{mohamed_self-supervised_2022}. The representations from these models have been shown to have strong task generalisability, enabling state-of-the-art performance on a range of speech tasks with only lightweight supervised downstream task heads \cite{yang_large-scale_2024}. However, these representations encode information about many variables affecting the speech signal in a distributed fashion, while speech tasks typically rely on only a fraction of the overall variability in the speech signal. For example, the output of a content task like \ac{asr} should not be affected by who is speaking, or the output of \ac{sid} by the words spoken.

Accordingly, a line of research aims to learn task-specific features, invariant to changes in nuisance variables. Contrastive learning, which encourages similar data samples to be close together in representation space and different data samples to be far apart, has seen use for learning speaker-invariant content representations \cite{qian_contentvec_2022}, and content-invariant speaker representations \cite{xia_self-supervised_2021, zhang_contrastive_2021}.

An alternative approach is learning \textit{disentangled representations} \cite{oord_neural_2017, lian_robust_2022, jiang_disentangled_2023}, where different sources of variability are separated in latent space \cite{bengio_representation_2013}. Early work in disentangled representation learning considered the problem of non-linear \ac{ica}, based on twin assumptions of independence between the variables of interest and Gaussianity \cite{kingma_auto-encoding_2013}. Unfortunately, these assumptions are unable to guarantee disentanglement \cite{hyvarinen_nonlinear_2023, hyvarinen_identifiability_2024}, with performance sensitive to random seed and architecture choices \cite{locatello_challenging_2019}.

Recent work in image processing has moved towards \textit{causal disentanglement} \cite{scholkopf_towards_2021, Yang_2021_CVPR}. These approaches replace the assumption that the variables of interest are independent with causally-related variables, and often make use of interventional data to provide disentanglement guarantees \cite{lippe_citris_2022, zhang_identifiability_2023, ahuja_interventional_2023, squires_linear_2023}. Such data consists of samples resulting from interventions on the causal variables which affect the data---the value of one or more variables is changed, and the causal effect is allowed to propagate through the system \cite{peters_elements_2018}.

In this work, we apply this idea to the post-training of speech foundation models. We propose a novel framework, \textit{interventional contrastive learning}, adapting supervised contrastive learning to jointly learn multiple task-specific subspaces, enabled by an interventional dataset with only weak labels. These labels accord with intervention targets---for a pair of samples, which causal variables are intervened on. In contrast to standard contrastive learning, this approach learns multiple embedding spaces, associated with different causal variables, rather than a single embedding space which is invariant to all but one variable of interest.

A training dataset is synthesised using a state-of-the-art zero-shot \ac{tts} system, allowing us to intervene on variables of \textit{content} (sentence) and \textit{speaker}. Using synthetic data has two advantages. First, it allows us to control the density of interventions present in the dataset---in the best case, every speaker saying every utterance. Second, an intervention is particularly intuitive in the synthetic case. Just as we can picture a real-world setting where interventions arise through an agent interacting with levers and buttons which are the inputs to a causal system, we can intervene on the output of the \ac{tts} system by controlling its inputs: a reference audio and a target text. %

The contributions of the work are as follows. First, we introduce a novel contrastive loss, described in \Secref{sec:loss}, which allows us to learn multiple subspaces associated with different causal variables. Second, we propose a synthetic interventional dataset, with dense interventions on content and speaker causal variables. This dataset is described in \Secref{sec:dataset}. Finally, we examine the separating capacity of our framework using a minimal architecture, and evaluation on speaker and content tasks, where success is measured by increased performance on the matched subspace and decreased performance on the mismatched subspace. The experimental setup is described in \Secref{sec:experiments}. \Secref{sec:results} discusses the results of our experiments, and the limitations of our approach.

\section{Interventional Contrastive Learning}
\label{sec:loss}

In contrast to standard supervised contrastive learning, which operates in a single embedding space, we learn multiple subspaces corresponding to distinct causal variables. In order to learn these subspaces, we leverage a dataset where, for each subspace, each sample shares a label with at least one---and often many---examples (positives), alongside multiple examples which do not share the same label (negatives). To formulate a contrastive loss which can handle multiple positives and multiple subspaces, we extend the multi-positive contrastive loss from \cite{khosla_supervised_2020, tian_stablerep_2023} to the problem of interventional contrastive learning.

Consider a batch of samples with embeddings $\{\vz^{(n)}\}_{n=1}^N$, where each embedding is partitioned into $J$ subspaces $\vz_1^{(n)},\dots,\vz_J^{(n)}$, each of which is associated with a causal variable $C_j$. For each subspace $j$, we consider the problem of matching an anchor sample $\va=\vz_j^{(i)}$, to a set of candidates $\{\vb_1,\vb_2,\dots,\vb_K\}$, where $\vb_k=\vz_j^{(k)}$ for the non-anchor samples in the batch $k=\{1,\dots,N\}\setminus i$. The likelihood that an anchor matches each candidate is given by the contrastive distribution $\vq$:

\begin{equation}
    \vq_{i,j} = \frac{\exp(\va\cdot\vb_i/\tau)}{\sum_{k=1}^K \exp(\va\cdot\vb_k/\tau)}
\end{equation}
Anchors and candidates are all $l_2$-normalised, and the temperature, $\tau$, is a hyper-parameter. A supervision signal is provided through the ground truth distribution $\vp$:

\begin{equation}
    \vp_{i,j} = \frac{\emM_{j,\va,\vb_i}}{\sum_{k=1}^K \emM_{j,\va,\vb_k}}
\end{equation}
where $\mM_j$ is a contrastive label for a batch, the binary elements of which describe which candidates match the anchor sample in subspace $j$.

The contrastive loss for subspace $j$ is the cross-entropy between ground-truth and contrastive distributions:

\begin{equation}
    \mathcal{L}_j^\text{con}=-\sum_{i=1}^N \vp_{i,j} \log \vq_{i,j}
\end{equation}

In order to directly penalise similarity between the subspaces, we adopt an orthogonality loss from \cite{bousmalis_domain_2016} as a regularising term. For a full batch of embeddings $\mZ$, the orthogonality loss between two subspaces $j$ and $k$ is

\begin{equation}
    \mathcal{L}^\text{orth}_{j,k} = ||\mZ_j^T\mZ_k||^2_F
\end{equation}
where $||\cdot||^2_F$ is the squared Frobenius norm.

The overall loss is a weighted sum of the contrastive losses for each subspace and the orthogonality losses between subspaces:

\begin{equation}
    \mathcal{L}=\lambda_\text{con}\sum_j \mathcal{L}^\text{con}_j + \lambda_\text{orth} \sum_j \sum_{k>j} \mathcal{L}^\text{orth}_{j,k}
\end{equation}
where $\lambda_\text{con}$ and $\lambda_\text{orth}$ are weights on the contrastive and orthogonality losses, respectively.

\begin{figure}
    \centering
    \input{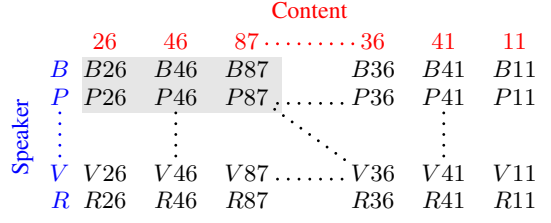}
    \caption{Example of construction of shuffled batches from the interventional dataset for a batch size of $6$. Speakers are labelled with letters, while content is labelled with numbers. The first batch is given by the shaded area.}
    \label{fig:batch_con}
\end{figure}

\section{Synthetic Interventional Dataset}
\label{sec:dataset}

In order to train a model with the objective described in \Secref{sec:loss}, we require an \textit{interventional dataset}, consisting of speech samples with labelled interventions on the causal variables of interest, $C$. Such a dataset could be constructed by mining large datasets of genuine speech for interventional pairs. However, we use a synthetic dataset in this work, which can have every combination of interventions on the causal variables. Specifically, we consider two causal variables: \textit{content} and \textit{speaker}. To obtain interventional data over these variables, we synthesise speech using a zero-shot \ac{tts} model, F5-TTS \cite{chen_f5-tts_2025}.

We source reference audio for zero-shot synthesis from the test-clean set of LibriTTS~\cite{zen_libritts_2019}. $32$ speakers are included in our training set, while $6$ make up the development set. For each speaker, we take a sample of $256$ reference utterances between $3$ and $10$ seconds in duration. Each of these is used as the reference audio for one synthesised utterance in the final dataset. 

The final dataset combines the $256$ reference utterances for each of the $38$ speakers with $256$ target texts for the appropriate subset, also sampled from LibriTTS. These are input to the \ac{tts} system to obtain synthesised utterances with exhaustive combinations of speakers and texts, such that the training set consists of $8192$ utterances, the development set of $1536$ utterances, and the two sets share neither speakers nor texts.

Batch construction involves arranging samples in a matrix, where each row is a speaker and each column is a specific content. Each batch is a section of this matrix, optimised such that the batches are as close to square as possible given the batch and dataset size. The rows and columns of the matrix are re-shuffled after every training epoch. Figure \ref{fig:batch_con} shows how this batch construction would work for batches of $6$ samples. For training, we use a batch size of $512$ with $16$ speakers and $32$ contents. The development set is split into batches of $768$ samples, such that each batch contains all $6$ speakers.

Interventional labels, $\mL_j$, indicate where content or speaker differs between two samples, and are defined for a batch. Contrastive labels, $\mM_j$, are then calculated from the interventional labels, such that:

\begin{equation}
    \mM_j = \mJ - \mL_j - \mI
\end{equation}
where $\mI$ is the identity matrix and $\mJ$ is the matrix of ones. Accordingly, we can map from our interventional labels to contrastive labels, which provide the supervision signal to the objective described in \Secref{sec:loss}.

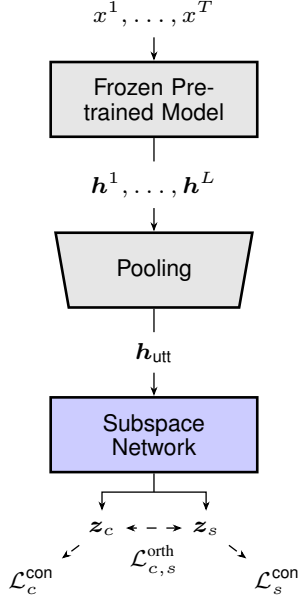
\begin{figure}
    \centering
    \resizebox{0.54\linewidth}{!}{
    \begin{tikzpicture}[node distance=0.25cm and 0.8cm]
    
        \tikzset{
          data_label/.style={text centered, font=\sffamily\scriptsize},
          base_box/.style={
              rectangle, draw, text width=2cm, minimum height=0.8cm, 
              text centered, font=\sffamily\scriptsize, thick
          },
          pooling/.style={
              trapezium, shape border rotate=180, trapezium angle=80, 
              draw, fill=gray!20, text width=0.8cm, minimum height=0.8cm, 
              text centered, font=\sffamily\scriptsize, thick
          },
          disentanglement/.style={base_box, fill=blue!20}, %
          frozen/.style={base_box, fill=gray!20},
          top_arrow/.style={draw},
          arrow/.style={draw, -{Stealth[length=1mm]}},
          dashed_line/.style={draw, dashed, -{Stealth[length=1mm]}},
          double_dashed_line/.style={draw, dashed, {Stealth[length=1mm]}-{Stealth[length=1mm]}}
        }
    
        \node (x) [data_label] {$x^1, \dots, x^T$};
        \node (pretrained) [frozen, below=of x] {Frozen Pretrained Model};
        \node (h) [data_label, below=of pretrained] {$\vh^1, \dots, \vh^L$};
        \node (pooling) [pooling, below=of h] {Pooling};
        \node (h_utt) [data_label, below=of pooling] {$\vh_{\text{utt}}$};
    
        \node (s) [disentanglement, below=of h_utt] {Subspace Network};
    
        \node (z_content) [data_label, below left=0.4cm and 0.3cm of s.south] {$\vz_c$};
        \node (z_speaker) [data_label, below right=0.4cm and 0.3cm of s.south] {$\vz_s$};
    
        \draw [arrow] (x) -- (pretrained);
        \draw [top_arrow] (pretrained) -- (h);
        \draw [arrow] (h) -- (pooling);
        \draw [top_arrow] (pooling) -- (h_utt);
        \draw [arrow] (h_utt) -- (s);
    
        \coordinate (s_branch) at ($(s.south) + (0, -0.2cm)$);
        \draw (s.south) -- (s_branch);
        \draw [arrow] (s_branch) -| (z_content.north);
        \draw [arrow] (s_branch) -| (z_speaker.north);
    
        \node (content_loss) [data_label, below left=0.2cm of z_content] {$\mathcal{L}^\text{con}_c$};
        \draw [dashed_line] (z_content) -- (content_loss);
        
        \node (speaker_loss) [data_label, below right=0.2cm of z_speaker] {$\mathcal{L}^\text{con}_s$};
        \draw [dashed_line] (z_speaker) -- (speaker_loss);
    
        \draw [double_dashed_line] (z_speaker) -- node[below, font=\scriptsize, yshift=-2pt] {$\mathcal{L}_{c,s}^{\text{orth}}$} (z_content);
        
    \end{tikzpicture}
}
    \caption{The architecture of the proposed model.}
    \label{fig:arch}
\end{figure}

\section{Experiments}
\label{sec:experiments}
We test our interventional contrastive post-training approach with the synthetic interventional dataset described in \Secref{sec:dataset} for a range of pre-trained models\footnote{https://github.com/mjukus/interventional-post-training-speech}.

For training, we use the AdamW optimiser and train for $50$ epochs. We use a one-cycle learning rate scheduler with cosine annealing, which performed better on the development set in initial experiments than constant or plateau-reduction schedulers. We used a maximum learning rate of $1\times10^{-4}$, starting percentage of $0.1$, division factor of $25.0$ and a final division factor of $1\times10^3$. All experiments were performed on a single Nvidia A100 GPU, where our models take $<2$~hours to train.

\subsection{Architecture}
\label{subsec:arch}
The model we apply to the problem of interventional contrastive learning consists of three sequential components. First, an utterance $\vx=(x^1, \dots, x^T)$ is transformed into a sequence of vectors $\mH=(\vh^1,\dots,\vh^L)$ by a frozen pretrained speech foundation model, or backbone. The length of this representation is proportional to the length of the original speech signal. Second, a pooling module aggregates the sequence into a single utterance embedding vector $\vh_\text{utt}$. Finally, a subspace projection network transforms this utterance embedding into a representation $\vz$, which is partitioned into subspaces, $\vz_j$, of equal size. Figure \ref{fig:arch} shows this architecture and the associated terms in the loss.

We conduct experiments with three popular speech foundation models as backbones. These models are wav2vec 2.0 Base \cite{baevski_wav2vec_2020}, HuBERT Base \cite{hsu_hubert_2021}, and WavLM Base \cite{chen_wavlm_2022}. All three models are based on the Transformer architecture, and are trained using masked quantised latents on the $960$~hours of LibriSpeech training data \cite{panayotov_librispeech_2015}. These models showed strong generalisability across tasks on the \mbox{SUPERB} benchmark \cite{yang_large-scale_2024}, suggesting that they encode both speaker and content information in their latent embeddings, and are therefore appropriate inputs for our approach.

We use the \mbox{SUPERB} implementations of the pre-trained models\footnote{https://github.com/s3prl/s3prl}, and extract features from the last layer of each model. This results in a sequence of vectors of dimension $768$, which are passed to the pooling module.

Based on work showing that mean-pooled representations from pre-trained speech foundation models perform well \cite{sanabria_analyzing_2023, duquenne_sonar_2023}, we use a mean-pooling approach with no trainable parameters. This enables direct evaluation of the effect of the subspace projection network, without introducing additional parameters.

The subspace projection network (henceforth subspace network) is a $3$-layer \ac{mlp}, consisting of an input layer, single hidden layer, and an output layer. This fully-connected network takes an utterance embedding vector of dimension $768$ and outputs a embedding $\vz$ of the same dimension. The hidden dimension is also $768$---we avoid a bottleneck dimension, because the aim of our approach is learning a transformation into a space of equal dimension. This is the smallest \ac{mlp} without a bottleneck, with a parameter count of $\sim 1.8$~M. The produced embedding vector is partitioned into a content subspace $\vz_c$ and a speaker subspace $\vz_s$, both of dimension $384$. The overall loss is composed of the contrastive loss for each of the two subspaces, and the orthogonality loss between the two subspaces.

\subsection{Baselines}
\label{subsec:baselines}
We compare our full architecture to a baseline that omits the subspace network. That is, the representations from this baseline are the pre-projection utterance embeddings $\vh_\text{utt}$ of our model, or the result of mean pooling on the appropriate backbone. This allows us to evaluate the benefits of the transformation to subspaces while controlling for all variables except the architecture and weights of the subspace network.

To quantify the benefit of jointly learning subspaces, we also compare our model with variants which learn only one subspace, which we refer to as the \textit{content only} and \textit{speaker only} models. These use the same contrastive loss, but are supervised only on interventional labels for one causal variable at a time. Accordingly, only one subspace is learned and there is no orthogonality term in the overall loss. To aid comparison with the full model, the dimension of the output from the subspace model is halved (to $384$), such that the capacity of the single subspace is the same as the equivalent subspace in the full model. These models have $~1.5 M$ parameters each.

\begin{table*}[ht]
    \centering
    \caption{All metrics are reported in \% as the mean (and standard error on the mean) over $5$ seeds, where applicable. The arrows $\uparrow$ and $\downarrow$ indicate whether a higher or lower value is better for a metric, respectively. Bold font indicates the best performing model for each backbone in the matched subspace; for the mismatched subspace, the worst performing model is bold instead. An asterisk indicates baseline results which have been repeated to aid comparison.}
  \begin{tabular}{c|c| c|c| c|c}
    \multirow{2}{*}{Backbone} & \multirow{2}{*}{\shortstack{Subspace\\Network}} & \multicolumn{2}{c|}{Matched Subspace} & \multicolumn{2}{c}{Mismatched Subspace} \\
     & & SV EER $\downarrow$ & KS acc $\uparrow$ & SV EER $\downarrow$ & KS acc $\uparrow$ \\
    \hline \hline
		\multirow{4}{*}{Wav2vec 2.0} & None & $45.5$ & $\mathbf{81.2~(0.6)}$ & $\mathbf{45.5}$* & $81.2~(0.6)$* \\
		 & Content only & - & $46.6~(0.4)$ & $44.0~(0.2)$ & - \\
		 & Speaker only & $37.3~(0.1)$ & - & - & $\mathbf{34.1~(0.4)}$ \\
		 & Full & $\mathbf{36.8~(0.2)}$ & $46.8~(0.5)$ & $43.2~(0.05)$ & $40.3~(0.6)$ \\
        \hline
		\multirow{4}{*}{HuBERT} & None & $36.4$ & $\mathbf{95.7~(0.1)}$ & $36.4$* & $95.7~(0.1)$* \\
		 & Content only & - & $90.7~(0.2)$ & $\mathbf{43.7~(0.1)}$ & - \\
		 & Speaker only & $\mathbf{26.9~(0.2)}$ & - & - & $\mathbf{83.2~(0.6)}$ \\
		 & Full & $\mathbf{27.2~(0.5)}$ & $89.7~(0.4)$ & $42.6~(0.2)$ & $85.1~(0.3)$ \\
        \hline
		\multirow{4}{*}{WavLM} & None & $38.7$ & $\mathbf{96.9~(0.04)}$ & $38.7$* & $96.9~(0.04)$* \\
		 & Content only & - & $93.8~(0.1)$ & $\mathbf{44.2~(0.1)}$ & - \\
		 & Speaker only & $\mathbf{24.6~(0.5)}$ & - & - & $\mathbf{81.5~(1.2)}$ \\
		 & Full & $\mathbf{24.7~(0.2)}$ & $93.0~(0.3)$ & $42.5~(0.2)$ & $84.9~(0.5)$ \\
        \hline
    \end{tabular}
    \label{tab:results}
\end{table*}

\subsection{Evaluation}
\label{subsec:eval}
In order to evaluate the information contained in the learned representations for each subspace, we use a cross-evaluation procedure on two utterance-level downstream tasks---one content task and one speaker task.

For the speaker variable, we evaluate both learnt subspaces on \ac{sv}. For a pair of enrollment and test utterances, \ac{sv} is a binary classification task, determining whether the speakers of the two utterances match. Following common practice in \ac{sv}, we report the \ac{eer} metric over the VoxCeleb1 test set \cite{nagrani_voxceleb_2020}. However, since we expect the speaker subspace learned by our subspace model to resemble a speaker embedding, we do not train a model for \ac{sv}. Instead, we consider the case of \ac{ood} \ac{sv}, calculating speaker similarity directly on the representations learnt by our subspace model, compared with the representations from the baselines. This is a significantly more challenging scenario---our synthetic data consists of clean read speech, while the VoxCeleb data is in-the-wild spontaneous speech. Success on this task relies on the generalising capacity of the backbone, and that the subspace model learns a data-agnostic transformation. The latter is a particularly desirable quality, because it enables us to learn a transformation on one dataset that is useful across many datasets and tasks.
For the content variable, we evaluate the learnt subspaces on \ac{ks}. This task involves recognising known keywords from other unknown content. We adopt the approach from SUPERB, which trains a downstream model consisting of a linear head and mean pooling on the Speech Commands dataset \cite{warden_speech_2018, yang_large-scale_2024}. The model is trained using the cross entropy loss to classify between $12$ classes, including classes for unknown words and non-speech. We use the Adam optimiser with a learning rate of $0.001$, training for $30$ epochs, and report \ac{acc} over the test set.

\section{Results}
\label{sec:results}
Table \ref{tab:results} shows the performance of each backbone and model combination on the \ac{ood} \ac{sv} and \ac{ks} tasks. Results are reported for both the matched subspace (where the representation corresponds to the target variable) and the mismatched subspace. 

For all backbones, the speaker subspace from the full model is substantially better for the \ac{ood} \ac{sv} task than without the subspace network. The absolute \ac{eer} for even the best models remains high at $\sim25$\%, when compared with results on the \mbox{SUPERB} \ac{sv} task---for example, an \ac{eer} of $4.69$\% on WavLM in \cite{yang_large-scale_2024}. However, this result used a $5$-layer time-delay neural network and the $1211$ speakers in the VoxCeleb1 training set, while we train a smaller model on data from just $32$ speakers. Furthermore, our data is clean, read speech, while VoxCeleb1 is in-the-wild, spontaneous speech. Despite the data mismatch and small training speaker pool, our best models successfully separate out speaker information from a pre-trained backbone, beating the \ac{eer} of $26.51$\% for non-expert humans reported in \cite{kang_augmentation_2022}.

On the other hand, our learned content subspace is less useful than the pooled backbone representations for \ac{ks}, indicating that content information is lost as a result of our contrastive objective. In order to analyse why this happens, compare the pretext task optimised in learning the content subspace with the \ac{ks} task. While the contrastive loss encourages similarity between utterances which are examples of the same sentence, \ac{ks} involves utterances which contain only single words. There is no guarantee that the word-level information useful for \ac{ks} and the utterance-level information useful for the pretext task coincide in the pre-trained embedding space. Accordingly, the transformation learnt by the pretext task may not prioritise word-level information. Nevertheless, the performance of our best models on \ac{ks} remains competitive with the baselines, if less so than for the speaker subspace.

The gap between \ac{sv} performance on the matched and mismatched subspaces provides evidence that the interventional contrastive loss does separate speaker and content information to some degree. The gap for \ac{ks} is less compelling, with absolute accuracy high for the HuBERT and WavLM backbones on both matched and mismatched subspaces. This indicates that either the \ac{ks} task is possible with limited content information, or that there is a substantial amount of information leakage between subspaces. That the models which train subspaces jointly generally perform better than the single subspace models in mismatched conditions provides further evidence of information leakage.

Across all backbones, there is no evidence of better performance from learning content and speaker subspaces jointly. At the same time, the performance of the single subspace and joint subspace models is similar, while the joint model has fewer total parameters than two single subspace models combined. That might suggest that the joint model uses its subspace network more efficiently, although further experiments would be needed to confirm this.

Finally, the models trained on wav2vec 2.0 see consistently worse performance. This is likely because we use the last layer of each model---it has been shown that the last layer of wav2vec 2.0 is generally less useful than earlier layers \cite{yang_large-scale_2024, mohamed24_interspeech}. Future work should experiment with different layers or a weighted sum of layers to test this.

\section{Conclusion}
We propose a novel post-training approach using a multi-part contrastive loss to learn separate content and speaker subspaces from interventional data. Our approach increases performance on \ac{ood} \ac{sv} and beats human performance, despite training on just $32$ speakers, while maintaining similar performance to the baselines on the \ac{ks} task, using just $256$ training utterances per speaker. However, results suggest that joint learning of the subspaces does not increase performance over learning the subspaces separately. Further work will look at how the approach scales to larger, real speech datasets, using different layers of the pre-trained backbones, and direct penalisation of information leakage in the loss.

\section{Acknowledgements}
\ifcameraready
     This work was supported by the UKRI AI Centre for Doctoral Training in Speech and Language Technologies (SLT) and their Applications funded by UK Research and Innovation [grant number EP/S023062/1] and by Meta. We also acknowledge IT Services at The University of Sheffield for the provision of services for High Performance Computing.
\else
     To maintain anonymity, acknowledgements are excluded from the review version of the manuscript.
\fi

\section{Generative AI Use Disclosure}
The authors used \textit{Gemini Pro} and \textit{Github Copilot (GPT-4.1)} for technical assistance with programming tasks, as well as in generation of prototypes for tables and figures in the manuscript. Generative AI was otherwise not used in the production of the manuscript.

\bibliographystyle{IEEEtran}
\bibliography{mybib}

\end{document}